\documentclass[runningheads]{llncs}

\usepackage[mobile]{eccv}

\usepackage{eccvabbrv}

\usepackage{graphicx}
\usepackage{booktabs}

\usepackage[accsupp]{axessibility}

\usepackage{hyperref}

\usepackage{orcidlink}

\title{\titletext}
\titlerunning{\method{}: Training-free, Open-Vocabulary Panoptic Occupancy Prediction via Foundation Models}

\author{Andrew Caunes\inst{1,2} \and
Thierry Chateau\inst{1} \and
Vincent Frémont\inst{2}}

\authorrunning{A. Caunes et al.}

\institute{Logiroad, Nantes, France \\
\and
LS2N - Ecole Centrale de Nantes, France\\
}

\usepackage{graphicx}
\usepackage{amsmath}
\usepackage{amssymb}

\usepackage[utf8]{inputenc}
\usepackage[T1]{fontenc}
\usepackage{lmodern}
\usepackage{url}
\usepackage{booktabs}
\usepackage{amsfonts}
\usepackage{nicefrac}
\usepackage{microtype}
\usepackage[dvipsnames]{xcolor}

\usepackage{amssymb}
\usepackage{pifont}
\usepackage{fixmetodonotes}
\usepackage{colortbl}
\usepackage{rotating}
\usepackage{scalerel}

\usepackage{multirow}
\usepackage[normalem]{ulem}
\usepackage{cancel}

\usepackage{bm}
\usepackage{adjustbox}
\usepackage{array}
\newcolumntype{R}[2]{
	>{\adjustbox{angle=#1,lap=1.3\width-(#2)}\bgroup}
	l
	<{\egroup}
}

\usepackage[linesnumbered,ruled,vlined]{algorithm2e}
\usepackage{algpseudocode}

\makeatletter
\@namedef{ver@everyshi.sty}{}
\makeatother
\usepackage{tikz}
\usetikzlibrary{positioning,shapes}

\definecolor{MyGreen}{RGB}{0, 180, 0}
\definecolor{MyRed}{RGB}{180, 0, 0}
\definecolor{MyBlue}{RGB}{30, 0, 180}
\definecolor{MyGrey}{RGB}{82.75, 82.75, 82.75}

\definecolor{skcar}{RGB}{100,150,245}
\definecolor{skbicycle}{RGB}{255, 200, 0}
\definecolor{skmotorcycle}{RGB}{255, 120, 0}
\definecolor{sktruck}{RGB}{80, 30, 180}
\definecolor{skotherv}{RGB}{0, 0, 255}
\definecolor{skpedestrian}{RGB}{255,30,30}
\definecolor{skdrivable}{RGB}{255, 0,255}
\definecolor{sksidewalk}{RGB}{75, 0,75}
\definecolor{skterrain}{RGB}{150, 240, 80}
\definecolor{skvegetation}{RGB}{0, 175, 0}
\definecolor{skbuilding}{RGB}{255, 200, 0}
\definecolor{mygreen}{RGB}{0, 170, 0}

\definecolor{tsne_source}{rgb}{0.86, 0.3712, 0.33999999999999997}
\definecolor{tsne_target}{rgb}{0.33999999999999997, 0.8287999999999999, 0.86}

\newcommand{\cmark}{{\textcolor{MyGreen}{\ding{51}}}}

\newcommand{\method}{{FreeOcc}}
\def \method{FreeOcc\xspace}

\usepackage{makecell}
\newcommand{\titletext}{\method{}: Training-free Panoptic Occupancy Prediction via Foundation Models}

\newcommand{\second}[1]{\cellcolor{blue!10}{#1}}
\newcommand{\best}[1]{\cellcolor{blue!20}{#1}}

\newcommand{\sem}{{\mathsf{sem}}}

\usepackage{stmaryrd}
\usepackage{trimclip}

\makeatletter
\DeclareRobustCommand{\shortto}{
	\mathrel{\mathpalette\short@to\relax}
}

\newcommand{\short@to}[2]{
	\mkern2mu
	\clipbox{{.5\width} 0 0 0}{$\m@th#1\vphantom{+}{\shortrightarrow}$}
}
\makeatother

\begin{document}
\maketitle
\begin{abstract}
	Semantic and panoptic occupancy prediction for road scene analysis 
	provides a dense 3D
	representation of the ego vehicle's surroundings.
	Current camera-only approaches typically rely on costly dense 3D 
	supervision or require training models on data from the target 
	domain, limiting deployment in unseen environments.
	We propose \method{}, a training-free pipeline that leverages 
	pretrained foundation models to recover both 
	semantics and geometry from multi-view images.
	\method{} extracts per-view panoptic priors with a promptable 
	foundation segmentation model and prompt-to-taxonomy rules, and reconstructs
	metric 3D points with a reconstruction foundation model.
	Depth- and confidence-aware filtering lifts reliable labels into 3D, 
	which are
	fused over time and voxelized with a deterministic refinement stack.
	For panoptic occupancy, instances are recovered by fitting and merging robust
	current-view 3D box candidates, enabling instance-aware occupancy without any
	learned 3D model.
	On Occ3D-nuScenes, \method{} achieves 16.9 mIoU and 16.5 RayIoU 
	train-free, on par with state-of-the-art weakly supervised methods. 
		When employed as a pseudo-label generation pipeline for training 
		downstream models, it achieves  21.1 RayIoU, surpassing 
	the previous state-of-the-art weakly supervised baseline.
	Furthermore, \method{} sets new baselines for both train-free and 
	weakly supervised panoptic occupancy prediction, achieving 
	3.1 RayPQ and 3.9 RayPQ, respectively.
	These results highlight foundation-model-driven perception as a practical 
	route to training-free 3D scene understanding. 
	\keywords{3D occupancy prediction \and Panoptic occupancy \and Foundation models \and Training-free inference \and Autonomous driving}
\end{abstract}

\section{Introduction}
\label{sec:introduction}
\begin{figure}
	\centering
	\IfFileExists{imgs/method_overview.pdf}{
		\includegraphics[width=\textwidth]{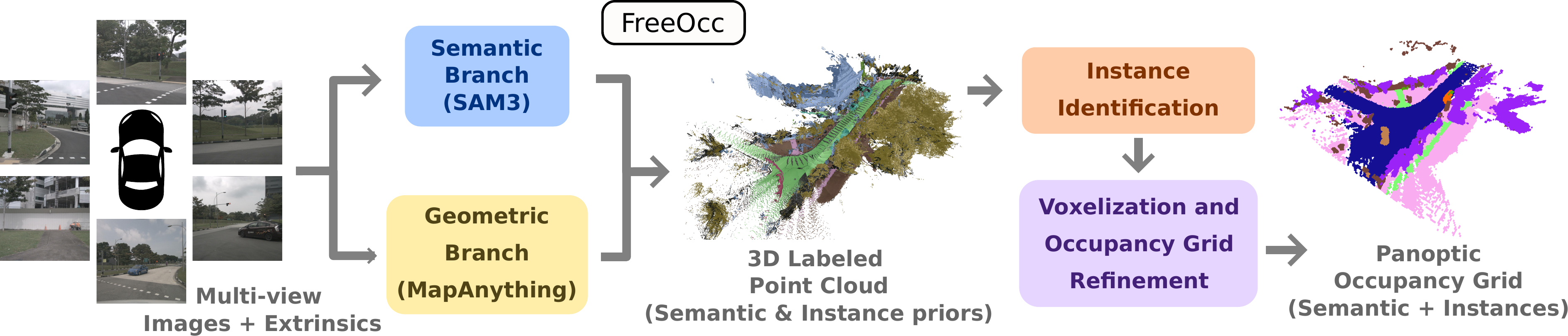}
	}{
		\fbox{\parbox[c][0.22\textheight][c]{0.9\textwidth}{
				\centering \textbf{Method overview figure (placeholder).}\\
				Add \texttt{figures/method\_overview.pdf}.
			}}
	}
	\caption{\textbf{\method{}: Training-free panoptic occupancy prediction from foundation models.}
	The pipeline operates directly
	from camera images and poses without any training. 
	It combines a Semantic Branch powered by the SegmentAnything 
	foundation model (SAM3) and a Geometric Branch powered by the
	MapAnything 3D reconstruction model to obtain 3D 
	semantically labeled point clouds with instance priors from SAM3's masks.
	An Instance Identification module then refines instance candidates and 
	re-assigns wrongly labeled points to the correct instance and semantic class.
	Finally, the fused point cloud is voxelized and refined to produce the final
	panoptic occupancy grid.
	}
	\label{fig:method_overview}
\end{figure}

Understanding and predicting the 3D structure of road scenes is a 
cornerstone of autonomous driving and road infrastructure analysis.
While active sensors such as LiDAR provide accurate geometry, they 
increase system cost and are not always available, motivating camera-only 
perception as a scalable alternative.

However, recovering metric 3D structure from RGB is inherently ambiguous: 
depth is only indirectly observable, and occlusions, long-range geometry, 
and dynamic actors further challenge dense 3D reasoning.

3D occupancy prediction addresses this by discretizing the ego-centric 
scene into a voxel grid and estimating which voxels are occupied, often 
with semantic labels; its panoptic variant additionally assigns instance 
identifiers to ``thing'' classes.
Benchmarks such as Occ3D \cite{tian_occ3d_2023} have standardized 
evaluation for camera-based semantic occupancy.

Most high-performing approaches rely on dense 3D supervision, 
typically generated from LiDAR annotation accumulation and post-processing.
Such supervision is costly to acquire and limits scaling to new domains, 
sensor configurations, and evolving taxonomies for semantic classes.

This has spurred weakly supervised alternatives that reduce manual 
3D labeling by leveraging geometric priors and pseudo-labels.
In particular, GaussianFlowOcc \cite{boeder_gaussianflowocc_2025} and 
ShelfOcc \cite{boeder_shelfocc_2025} demonstrate that foundation-model-driven 
cues can provide useful supervision for occupancy learning, yet they still 
depend on training a target-domain occupancy network and focus primarily 
on semantic occupancy.

We propose \method{}, a foundation-model pipeline that pushes this 
paradigm to the limit: \method{} performs \emph{training-free} semantic
\emph{and} panoptic occupancy prediction 
directly at inference time.
It requires no target-domain images prior to deployment, can be run in a 
new domain without any optimization, and inherits the open-vocabulary 
capabilities of foundation models, allowing it to adapt its label space 
by changing text prompts rather than retraining a 3D model.

The main cost of eliminating training is compute: running large 2D and 3D 
foundation models makes inference slower than a specialized network.
We therefore position \method{} first as a stand-alone train-free
predictor for rapid deployment that can alternatively
be used as a high-quality pseudo-label generator 
for training downstream real-time occupancy models when target data 
becomes available.

Concretely, \method{} combines a promptable segmentation foundation model 
(SAM \cite{kirillov_segment_2023}; we use SAM3) to extract per-view 
semantic/instance masks with confidences, and a metric reconstruction model 
(MapAnything \cite{keetha_mapanything_nodate}) that outputs dense per-pixel 
3D points along with depth and confidence maps.
Depending on the application, this pipeline can be applied both in a causal 
and non-causal manner, by either only using past and current frames, or all 
available frames.
After confidence-aware filtering, reliable per-pixel 3D points are assigned 
semantic and instance labels, fused over time, and converted into a voxel grid 
with a deterministic multi-stage refinement stack.
For panoptic occupancy, we fit and merge robust 3D box candidates from 
current-view thing evidence to assign consistent instance ids in 
the fused cloud prior to voxelization.

On Occ3D-nuScenes \cite{tian_occ3d_2023}, \method{} achieves 16.9 mIoU 
and 16.5 RayIoU for semantic occupancy without any 
training (Table~\ref{tab:train_free}), substantially improving over 
prior train-free results (e.g., ShelfOcc at 9.6 mIoU \cite{boeder_shelfocc_2025}).
Despite being train-free, our semantic results are competitive with trained 
weakly supervised occupancy methods such as 
GaussianFlowOcc \cite{boeder_gaussianflowocc_2025}.
For comparison purposes, we also train downstream occupancy models on 
pseudo-labels generated by \method{}, following \cite{boeder_shelfocc_2025},
and reach state-of-the-art weakly supervised performance with 21.1 RayIoU.
For panoptic occupancy, we report the first training-free and 
weakly supervised results on 
Occ3D-nuScenes, with a training-free baseline at 3.1 RayPQ and a
weakly supervised transfer result at 3.9 RayPQ, highlighting both the 
feasibility 
of instance-aware label-free occupancy and the remaining gap to fully 
supervised predictors \cite{liu_fully_2024,yu_panoptic-flashocc_2024}.
We further provide an ablation study and detailed analysis of the impact 
of each pipeline component.
\textbf{Our main contributions are the following:}

\begin{itemize}
	\item

	      \textbf{Training-free.} We introduce \method{}, a 
		  training-free occupancy prediction pipeline that matches or surpasses 
		  trained weakly supervised baselines.
		  A multi-stage refinement of foundation-model 
		  outputs allows it to reach 16.9 mIoU and 16.5 RayIoU for semantic occupancy
		  on Occ3D-nuScenes without any training or target data.
		\item
		  \textbf{Pseudo-label generation.}
		  It can also be used as a pseudo-label generator for training downstream 
		  occupancy models for real-time inference. In this setting, it 
		  achieves state-of-the-art performance:
		  22.8 mIoU and a 21.1 RayIoU for semantic occupancy,
		  without the use of visibility masks during training.

	\item

	      \textbf{Panoptic occupancy.} Using an efficient
		  instance identification module, \method{} can also output 
		  robust instance predictions. It achieves
		  3.1 RayPQ (train-free) and 3.9 RayPQ (weakly supervised) 
		  on Occ3D-nuScenes, setting new baselines for both regimes.
\end{itemize}

\section{Related works}
\label{sec:related}

\subsection{3D Occupancy Prediction}

Camera-based 3D occupancy prediction estimates a voxelized representation 
of the scene from multi-view images and calibrated poses, with semantic 
labels assigned to occupied voxels.
Voxels in a fixed grid are more straightforward to predict from 2D images
than unstructured point clouds, making this the natural setting for camera-only 
prediction methods.
Panoptic occupancy extends this setting by adding instance 
identifiers for ``thing'' classes, combining volumetric semantics 
and instance-level scene decomposition in a single output.
Occ3D \cite{tian_occ3d_2023} standardized the semantic occupancy prediction 
task at scale, and 
later works further emphasized panoptic occupancy prediction 
with ray-based evaluation metrics for better 
geometric fidelity analysis \cite{liu_fully_2024}.

Most high-performing camera-only methods on Occ3D are trained with
dense 3D voxel labels.
Representative architectures include dense volumetric lifting and 3D
refinement pipelines \cite{wei_surroundocc_2023}, and sparse formulations
that avoid dense 3D reasoning via sparse reconstruction and query-based
prediction \cite{liu_fully_2024}.
In parallel, several works explore alternative 3D feature parameterizations,
e.g., multi-plane (tri-plane/TPV-style) representations \cite{huang_tri-perspective_2023},
which have been adopted by later occupancy systems.
More recent systems revisit the output space itself, e.g., set prediction
formulations \cite{wang_opus_nodate}.
Temporal reasoning is a major performance driver, with STCOcc \cite{liao_stcocc_2025}
showing strong gains via explicit sparse spatial-temporal cascade renovation.

Panoptic occupancy prediction is harder and has been less explored 
than semantic occupancy as it requires
both 3D geometric consistency and instance-aware grouping in a voxelized space.
SparseOcc \cite{liu_fully_2024} is a strong fully-supervised baseline and
formalized a panoptic evaluation benchmark with the ray-based RayPQ metric.
Panoptic-FlashOcc \cite{yu_panoptic-flashocc_2024} targets efficiency with a
streamlined design that combines semantic occupancy with instance-center
prediction and lightweight panoptic assignment.

\subsection{Weakly Supervised / Training-Free Paradigms}

To reduce reliance on dense 3D voxel supervision, prior works explored several
regimes that trade 3D GT for weaker signals.

Some methods replace 3D voxel GT with supervision available in the image plane
(e.g., semantic labels/depth maps), typically implemented via differentiable
rendering/projection losses \cite{pan_renderocc_2024}. While avoiding dense 3D
annotations, these approaches still rely on labeled 2D datasets.

A related line keeps the same 2D training paradigm but replaces human 2D labels by
pseudo-labels produced by off-the-shelf models (e.g., 2D semantics/depth) 
and/or self-supervised multi-view/video consistency losses. 
This includes GaussianFlowOcc \cite{boeder_gaussianflowocc_2025} 
as well as NeRF-/Gaussian-based variants and foundation-model alignment schemes
\cite{zhang_occnerf_2024,gan_gaussianocc_2025,jiang_gausstr_2025}.

Most closely related to our setting, two recent preprints, 
EasyOcc \cite{hayes_easyocc_2025} and ShelfOcc \cite{boeder_shelfocc_2025} 
proposed to leverage foundation-model priors to generate
3D voxel pseudo-labels, which are then used to train a
downstream occupancy network. For fairness, we include and compare against
them in our experiments using their reported results.
ShelfOcc \cite{boeder_shelfocc_2025} in particular
leverages off-the-shelf geometric and semantic foundation models to build
metrically consistent 3D voxel pseudo-labels, which are then used to train a
downstream model.
Our approach also leverages similar foundation models to obtain 
3D occupancy predictions. Differently from ShelfOcc, \method{}
is not pseudo-label focused.
Instead, we introduce multiple innovations to refine predictions 
such as a semantic prompt \& rules system, 
a multi-stage occupancy grid refinement module, 
and an instance identification module to re-assign 
instance and semantic labels.
These innovations allow our method to reach competitive
performance with ShelfOcc without their training scheme.

In contrast to all trained pipelines above, \method{} operates directly
at inference time without any downstream training. This enables deployment to
new environments without a separate data-collection and training phase. 
This setting also better preserves the open-vocabulary potential of 
foundation models as target 
classes can be picked on-the-fly at inference time.
When real-time inference is required, our pipeline can still be used as a 
pseudo-label generator for downstream training. We demonstrate this 
in our experiments for direct comparison with 
ShelfOcc \cite{boeder_shelfocc_2025} and 
EasyOcc \cite{hayes_easyocc_2025}.

Finally, existing weakly supervised works have focused on semantic 
occupancy prediction; to our knowledge, our method is the first to establish 
baselines for both weakly supervised and training-free panoptic occupancy 
prediction.

\section{Method}
\label{sec:method}
An overview of \method{} can be seen in Fig.~\ref{fig:method_overview}.
Its main stages are illustrated in
Figs.~\ref{fig:method_semantic_branch}--\ref{fig:method_voxelization} and 
explained in the following subsections.

\subsection{Problem Setup}
We consider multi-camera samples $\{\mathcal{X}_t\}_{t=1}^{T}$ with
$\mathcal{X}_t=\{(I_t^c,K_t^c,T_t^c)\}_{c=1}^{C}$,
where $I_t^c$ is the image from camera $c$, and $K_t^c,T_t^c$ are the 
corresponding intrinsics and extrinsics.

Our output at time $t$ is a semantic occupancy grid $\mathbf{Y}^{\sem}_t$ 
and an instance-id grid $\mathbf{Y}^{\mathrm{ins}}_t$ defined on a fixed 
ego-centric voxel lattice.
The pipeline is training-free and can be run either non-causally (offline) 
or causally; causality is enforced by restricting which frames are used to 
build the 3D evidence for time $t$.
We denote by $\mathcal{W}_t\subseteq\{1,\dots,T\}$ the set of frame indices
used to build 3D evidence for time $t$.
In the causal setting, $\mathcal{W}_t\subseteq\{1,\dots,t\}$.

\subsection{Prompted 2D Priors from SAM3}
\begin{figure*}[t]
	\centering
	\IfFileExists{imgs/method_semantic_branch.pdf}{
		\includegraphics[width=0.8\textwidth]{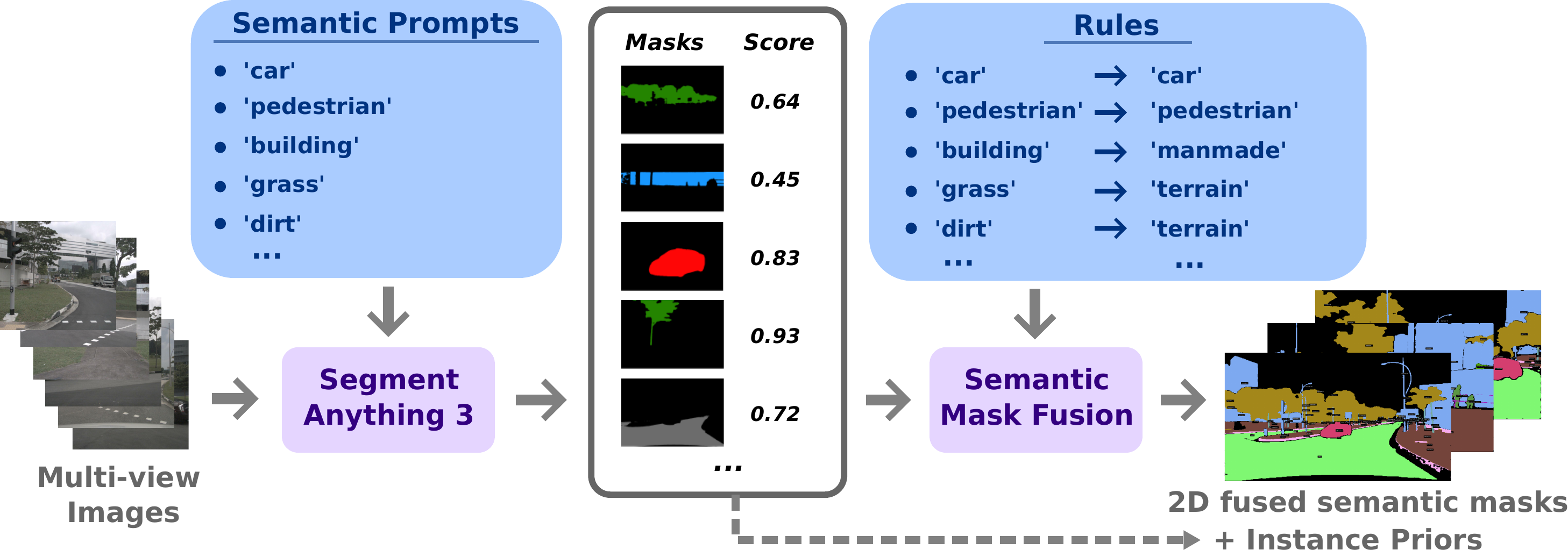}
	}{
		\fbox{\parbox[c][0.22\textheight][c]{0.9\textwidth}{
				\centering \textbf{Method overview figure (placeholder).}\\
				Add \texttt{figures/method\_semantic_branch.pdf}.
			}}
	}
	\caption{\textbf{Semantic branch.}
	A set of easily handcrafted prompts is fed to 
	SegmentAnything3 \cite{kirillov_segment_2023}, which yields 
	2D mask candidates with scores; we fuse 
	them into per-view semantic and instance priors and use rules 
	to remap prompt classes to the target taxonomy. Multiple prompts can 
	be used for each target class, e.g. synonyms.
	}
	\label{fig:method_semantic_branch}

\end{figure*}
The semantic branch details are illustrated in
Fig.~\ref{fig:method_semantic_branch}.
\textbf{Semantic prompts.}
We prompt a segmentation foundation model (SegmentAnything~\cite{kirillov_segment_2023}; 
we use SAM3) with a handcrafted prompt set $\mathcal{P}$, which is easily derived 
from the target taxonomy labels. This set includes synonyms intended 
to be more common
for the model than the class name. For example, for the class ``terrain'', we 
use the prompts ``grass'' and ``dirt'', yielding high-quality masks, while the ``terrain''
prompt is misunderstood. Similarly, the ``manmade'' class is too vague, while 
``building'' and ``wall'' are more specific and yield better results. 
We use 25 prompts in total for the 15 Occ3D classes 
(excluding 'others' and 'other\_flat').
The prompt set 
we use for the Occ3D-nuScenes taxonomy is provided in the Supplementary Material.
We show how naive class-name prompts 
affect performance significantly in the ablation study
in Sec.~\ref{sec:experiments}.

\textbf{Fusing masks.}
For each view, SAM3 produces multiple mask candidates per prompt, 
each with a score.
In this subsection, we omit the view index $(t,c)$ for clarity: 
all quantities are per-view.
For each prompt $k$ and candidate index $m$, SAM3 outputs a 
binary mask $M_{k}^{(m)}$ over the image domain with 
confidence score $s_{k}^{(m)}$.
We fuse candidates into a single semantic mask $\hat S(u)$ and an
instance prior mask $\hat U(u)$ by keeping the highest-scoring candidate 
covering each pixel:
\begin{equation}
	(\hat k,\hat m)(u)=\arg\max_{k,m:\,u\in M_{k}^{(m)}} s_{k}^{(m)}.
\end{equation}

\textbf{Applying rules.}
Finally, we remap the winning prompt label to the target taxonomy with 
simple mapping rules $r$:
\begin{equation}
	\hat S(u)=r(\hat k(u)),\qquad \hat U(u)=\hat m(u).
\end{equation}
This remapping collapses fine-grained prompts into target classes; 
for example, ``grass''$\rightarrow$``terrain'' and 
``building''$\rightarrow$``manmade''.

For the Occ3D-nuScenes taxonomy, the rule system is purely a 
prompt-to-taxonomy mapping, but there may also be cases of conflicts 
between classes, e.g.\ if the target taxonomy contains ``road'' and ``lane marking'', 
SAM3 may include lane markings in the ``road'' masks. In addition, the confidence 
score for ``road'' may tend to be higher than for ``lane marking'' because of the 
larger area. In these cases, rules can also incorporate 
``over''/``under'' relations between 
classes that will override the confidence score priority.

\begin{figure*}[t]
	\centering
	\IfFileExists{imgs/method_geometric_branch.pdf}{
		\includegraphics[width=\textwidth]{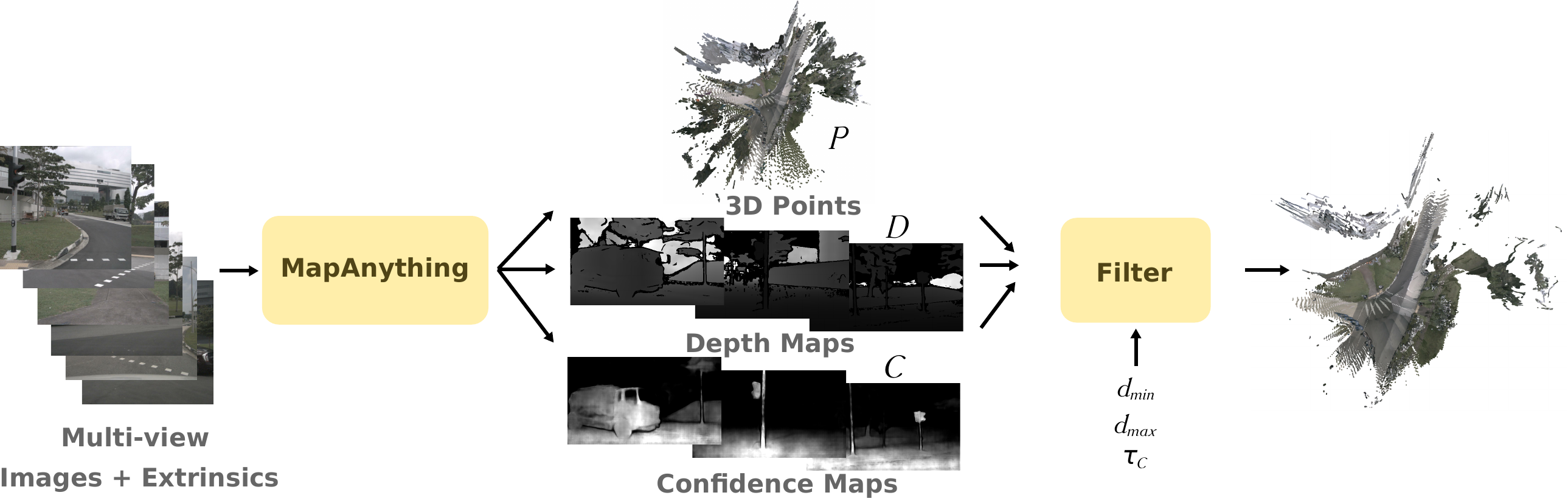}
	}{
		\fbox{\parbox[c][0.22\textheight][c]{0.9\textwidth}{
				\centering \textbf{Method overview figure (placeholder).}\\
				Add \texttt{figures/method\_geometric_branch.pdf}.
			}}
	}
	\caption{\textbf{geometric branch.}
		In parallel, MapAnything outputs per-pixel 3D points along with depth and confidence maps.
		We filter points by depth/confidence and keep the remaining labeled 
		3D points, yielding a sparse point cloud.
	}
	\label{fig:method_geometric_branch}
\end{figure*}
\subsection{Metric 3D Reconstruction and Reliability Filtering}
The geometric branch details are illustrated in
Fig.~\ref{fig:method_geometric_branch}.

A pretrained geometry model (MapAnything~\cite{keetha_mapanything_nodate}) 
predicts per-view 3D points, depth and confidence:
\begin{equation}
	(P,D,C)=\mathcal{G}(I,K,T).
\end{equation}

For filtering, confidence is stabilized by replacing non-finite values 
and applying log scaling:
\begin{equation}
	\tilde C(u)=
	\begin{cases}
		\log_{10}(C(u))+1, & C(u)\in\mathbb{R}_+\ \text{finite}, \\
		1,                     & \text{otherwise},
	\end{cases}
\end{equation}
We normalize confidence and use distance thresholds $d_{\min},d_{\max}$ 
and confidence threshold $\tau_C$ to filter unreliable points:
\begin{equation}
	\Omega=\{u:\, \tilde C(u)\ge \tau_C,\ d_{\min}\le D(u)\le d_{\max}\}.
\end{equation}

Each valid pixel yields a labeled 3D point, which 
inherits $(\hat S(u),\hat U(u))$ as its semantic label and instance prior
from the semantic branch.
This yields labeled point sets for all frames considered in $\mathcal{W}_t$.

\subsection{Instance Identification via Current-Sample 3D Candidates}
\begin{figure*}[t]
	\centering
	\IfFileExists{imgs/method_instance_identification.pdf}{
		\includegraphics[width=\textwidth]{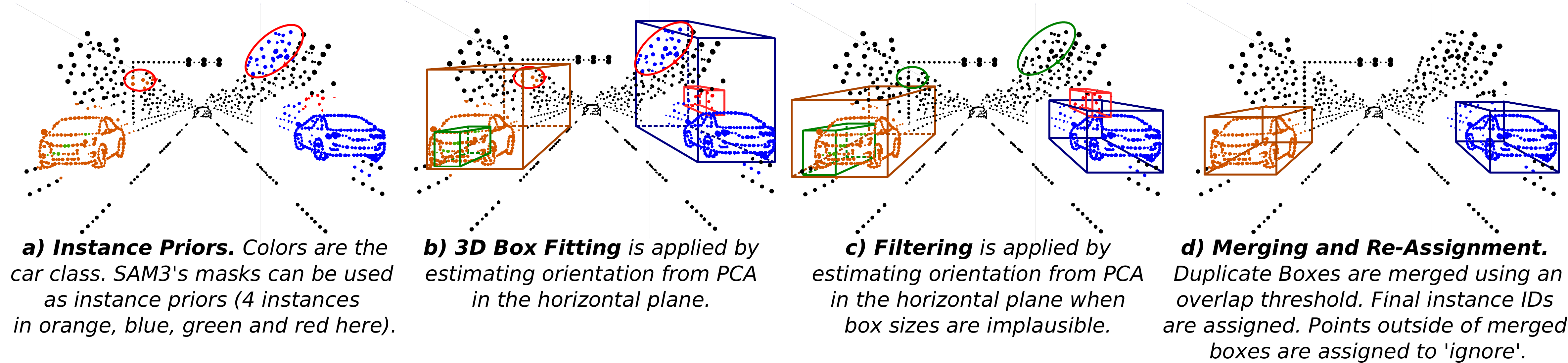}
	}{
		\fbox{\parbox[c][0.22\textheight][c]{\textwidth}{
				\centering \textbf{Method overview figure (placeholder).}\\
			}}
	}
	\caption{\textbf{instance identification.}
	Prompted SAM3 yields 2D mask candidates with scores; we fuse them into per-view semantic and instance priors and remap prompts to a canonical taxonomy with a simple lookup.
	MapAnything yields metric depth with a confidence signal; after reliability filtering we lift pixels to labeled 3D points.
	We optionally fuse points over time (non-causal or causal by frame selection), regularize thing instances via current-sample 3D box candidates, then voxelize and refine to produce semantic and panoptic occupancy grids.}
	\label{fig:method_instance_identification}
\end{figure*}

This stage is visualized in Fig.~\ref{fig:method_instance_identification}.
Temporal fusion improves static structure but can create ghosting when dynamic 
thing instances move. We therefore identify instances using 
current-sample evidence only. 
Instance priors derived from SAM3's masks are good candidates for 
final instances. Noise can however appear from overlap between the multiple
cameras and projection bleeding due to imprecise masks / extrinsics. We use an instance
identification module to filter out such noise and derive final instances. This
module can also improve semantic performance by de-assigning outliers.

\textbf{3D Box Fitting.}
We start with fitting yaw-oriented 3D boxes to all current-sample instance priors
(3D points designated by a single mask). 
We estimate the ground-plane orientation from PCA in 
the horizontal plane and take tight extents in the rotated frame 
($SE(2)$-like yaw + translation).

\textbf{Filtering.}
For each candidate box, we compare its size with class-wise 
plausible size intervals, e.g. a 5m area is implausible for a 
``pedestrian'' class. Exact intervals are given in the Supplementary Material. 
We then discard depth outliers 
with an 
interquartile-range (IQR) rule (optionally per camera/view), and prune 
geometric outliers using a PCA-based robust deviation test in the projected 
space. 
When a candidate still produces an implausibly large box, we iteratively 
tighten these robust thresholds and refit for a few iterations (max. 4) 
until the box becomes coherent or we reach a maximum number of passes. 

\textbf{Merging.}
Same-class candidate boxes are then merged 
conservatively to avoid duplicates, based on 
3D intersection-over-smaller-volume (IoSV) overlap: we merge boxes 
only when a large fraction of the smaller box is covered by the other, 
controlled by threshold $\tau_{ov}$. 

\textbf{Re-assignment.}
Finally, we assign one final instance id per 
merged box and reassign fused thing points: points inside a box take its id; 
uncovered points are assigned to the nearest box if their point-to-box 
distance is below $d_{nn}=2$m; otherwise they are relabeled as 'ignore'. 
This produces a fused cloud with more consistent thing instances prior to 
voxelization.
Stuff voxels receive canonical class-level ids.

\subsection{Occupancy Grids and Refinement}
\begin{figure*}[t]
	\centering
	\IfFileExists{imgs/method_voxelization.pdf}{
		\includegraphics[width=\textwidth]{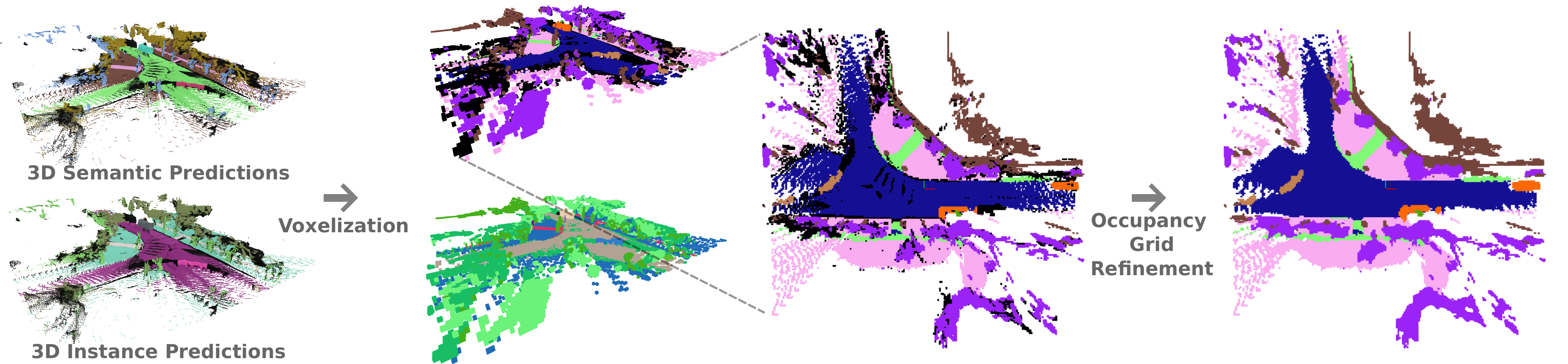}
	}{
		\fbox{\parbox[c][0.22\textheight][c]{0.9\textwidth}{
				\centering \textbf{Method overview figure (placeholder).}\\
				Add \texttt{figures/method\_voxelization.pdf}.
			}}
	}
	\caption{\textbf{voxelization branch.}
	We voxelize the labeled point cloud and apply a lightweight multi-stage 
	refinement step to produce semantic and panoptic occupancy grids. Notice
	that small holes, ego vehicle noise and backgrounded ghost artifacts 
	are removed in this stage.
	}
	\label{fig:method_voxelization}
\end{figure*}
The voxelization and refinement stages are illustrated in
Fig.~\ref{fig:method_voxelization}.
We crop the fused cloud to the occupancy-grid bounds and voxelize it with voxel
size $s_v$ on a grid of size $(X,Y,Z)$ with vertical offset $z_0$.

\textbf{Voxelization by voting.}
Each point is mapped to a voxel index, and each voxel is assigned a semantic
label by majority voting over the point labels while excluding 'ignore' votes.
Voxels with no non-ignore votes are assigned the 'free' class.
We keep voxels even with very few points ($n_{\min}=1$), as noisy returns are
already reduced by the per-view minimum-distance filtering applied after
reconstruction.

\textbf{Voxel evidence: support and confidence.}
For each occupied voxel $v$, we record its \emph{point support} $n(v)$, i.e.,
the number of points that voted for it, and we compute a smoothed vote
confidence for the winning class $c^*(v)$:
\[
\mathrm{conf}(v)=\frac{n_{c^*}(v)+\alpha}{n(v)+\alpha\,K},
\]
where $n_{c^*}(v)$ is the winning vote count, $K$ is the number of semantic
classes, and $\alpha$ is a small Dirichlet smoothing constant (we use
$\alpha=0.5$).
We also convert point support into a saturating reliability score
$p_{\mathrm{occ}}(v)=1-\exp(-\lambda\,n(v))$ (with $\lambda=0.35$), which
quickly approaches $1$ for high-support voxels.

\textbf{Refinement stack.}
We apply a deterministic four-stage refinement that improves local consistency
without over-smoothing.
Let $\mathcal{N}_{26}(v)$ denote the $3{\times}3{\times}3$ neighborhood around
$v$ (excluding $v$). The \emph{modal class} is the most frequent semantic label
in $\mathcal{N}_{26}(v)$, and the \emph{modal support} is its vote count (number
of neighbors carrying that label). We also denote by $n_{\mathrm{occ}}(v)$ the
number of non-unknown neighbors.

\textbf{(1) Pinhole and cavity filling.}
We close tiny empty holes inside locally occupied regions using 3D morphological
closing of the occupied mask.
Newly closed voxels that were previously free/ignore are filled with the modal class
when modal support is strong (at least $4$ agreeing neighbors).
To fill slightly larger semantic cavities, we use a stricter criterion and only
fill free/ignore voxels when the neighborhood is dense
($n_{\mathrm{occ}}(v)\ge 10$) and the modal support is high (at least $5$ votes).

\textbf{(2) Warmup ego completion.}
When temporal evidence is limited in early causal frames, we optionally
complete the near-ego blind region by filling unknown near-ground voxels as
'driveable surface'.
We consider voxels within radius $r_{\text{ego}}$ around the ego

and fill them as driveable only if they are close to
existing driveable evidence (via a small planar dilation) and not adjacent to
object voxels (via a small 3D dilation), to avoid hallucinating structures near
vehicles or pedestrians.

\textbf{(3) Conservative neighborhood coherence.}
We perform a single coherence pass on occupied voxels.
Voxels are \emph{frozen} if they are reliable according to voxel evidence
(e.g., $\mathrm{conf}(v)\ge 0.75$ or $p_{\mathrm{occ}}(v)\ge 0.85$), and we also
protect object and thin classes from being overwritten.
For the remaining ambiguous occupied voxels, we update the label to the modal
class only when neighborhood agreement is both strong and dominant (modal support
$\ge 5$ and modal votes form a clear majority, e.g.,
$\mathrm{support}_{\text{mode}}(v)\ge 0.6\,n_{\mathrm{occ}}(v)$).

\textbf{(4) Background cleanup and instance dilation.}
Finally, any residual ignore-labeled voxels are reassigned to the modal class
when supported (at least $2$ agreeing neighbors).
To reduce small unlabeled gaps in thing instances, we apply a class-constrained
instance dilation of radius $\delta=2$ voxels: within each thing semantic region,
unassigned instance voxels inherit the nearest instance id (in 3D Euclidean
distance) from nearby assigned voxels, improving instance completeness under
occlusions and sparse sampling.

\section{Experiments}
\label{sec:experiments}

We study \method{} under three complementary regimes.
First, we evaluate the strict train-free setting, which is the main contribution
of this work: no target-domain occupancy model is trained, and predictions are
produced directly from foundation-model inference and geometric fusion.
Second, for comparison with prior pseudo-label 
pipelines in the weakly supervised setting, we compare our train-free results, but also
train downstream STCOcc models from generated pseudo-labels.
Third, we report panoptic occupancy prediction performance, thus 
establishing baselines for train-free and weakly supervised panoptic 
occupancy prediction.
Finally, we perform module-level ablations to analyze where
the gains come from and which parts of the pipeline matter most.

We evaluate on Occ3D-nuScenes \cite{tian_occ3d_2023}.
Regrettably, no official test set is available for this dataset.
We follow the standard practice and report results on the 150 scene 
validation split.
For panoptic occupancy, we follow the SparseOcc protocol
\cite{liu_fully_2024} to generate instance voxel labels from nuScenes 
\cite{caesar_nuscenes_2020}.
For semantic occupancy, we report mIoU, IoU$_{occ}$, and RayIoU
(including RayIoU$_{1m}$, RayIoU$_{2m}$, and RayIoU$_{4m}$):
mIoU captures voxel-level semantic overlap, IoU$_{occ}$ captures binary occupancy,
and RayIoU more robustly evaluates semantic performance.
For panoptic occupancy, we report RayPQ and its range-specific variants
as defined in 
\cite{liu_fully_2024}, 
which jointly penalize semantic,
instance, and geometric errors.
To guarantee reproducibility, we directly use the official evaluation code
from SparseOcc \cite{liu_fully_2024} for all metrics.

\subsection{Implementation Details}

Unless stated otherwise, all reported train-free results are causal, i.e. only
past and current frames are used for each timestamp.
All hyperparameters are fixed and reused across runs after 
tuning
on a set of 30 training scenes:
$\tau_C=1e{-5}$, $d_{\min}=1$, $d_{\max}=50$, merging threshold
$\tau_{ov}=0.45$ (Intersection-over-Smaller-Volume),
grid bounds $(X,Y,Z)=(-40,40)\times(-40,40)\times(-10,10)$,
voxel size $s_v=0.4$m, vertical offset $z_0=-1$m, and dilation radius
$\delta=2$.
We do not retune parameters per comparison method or per metric.
Full implementation details will be made publicly available.

For weakly supervised comparison, we additionally train downstream STCOcc
models \cite{liao_stcocc_2025} from pseudo-labels generated by our pipeline in
non-causal mode (future frames contributing to current prediction).
We keep the official STCOcc training protocol (2 GPUs, batch size 8, 12
epochs) for the weakly supervised experiment.

For the panoptic occupancy prediction experiment, 
we extend STCOcc to jointly predict voxel semantics and instance IDs. 
The resulting model denoted STCOcc\_pano, inspired by SparseOcc's instance 
prediction, reuses STCOcc’s sparse 
3D BEV/voxel backbone and adds an instance-aware voxel head, panoptic label 
loading, and instance supervision (center focal-BCE + offset L1).

\subsection{Main Results}

\begin{table}
	\centering
	\caption{Training-free semantic occupancy prediction performance on 
	Occ3D-nuScenes validation set. IoU per class and mIoU are reported.
	'others' and 'other\_flat' classes are not included in the mIoU calculation.
	* means that the method is from a preprint.}
	\newcommand{\rotext}[1]{{\begin{turn}{90}{#1}\end{turn}}}

	\resizebox{\textwidth}{!}{

		\begin{tabular}{l|ccccccccccccccccc}
			\hline
			Method                                & \rotext{mIoU} & \rotext{occupied} & \rotext{barrier} & \rotext{bicycle} & \rotext{bus}  & \rotext{car}  & \rotext{cons. veh.} & \rotext{motorcycle} & \rotext{pedestrian} & \rotext{traffic cone} & \rotext{trailer} & \rotext{truck} & \rotext{driv. surf.} & \rotext{sidewalk} & \rotext{terrain} & \rotext{manmade} & \rotext{vegetation} \\
			\midrule

			ShelfOcc* \cite{boeder_shelfocc_2025} & 9.6           & 26.0              & 6.6              & 3.3              & 7.0           & 8.8           & 2.6                 & 4.7                 & 5.0                 & \textbf{8.1}          & 0.1              & 5.4            & 34.6                 & 14.6              & 18.2             & 10.9             & 14.5               \\

			\hline

			\method{} (Ours)                                  & \textbf{16.9} & \textbf{37.2}     & \textbf{16.1}    & \textbf{7.0}     & \textbf{12.2} & \textbf{17.6} & \textbf{13.0}       & \textbf{8.4}        & \textbf{9.7}        & 4.6                   & \textbf{2.1}     & \textbf{14.0}  & \textbf{51.3}        & \textbf{31.1}     & \textbf{29.0}    & \textbf{17.7}    & \textbf{19.4}      \\

			\bottomrule
		\end{tabular}
	}
	\label{tab:train_free}
\end{table}

Table~\ref{tab:train_free} compares IoU results in the train-free setting.
The only comparable baseline is ShelfOcc without downstream training
\cite{boeder_shelfocc_2025}. \method{} improves mIoU from 9.6 to 16.9
(+7.3 points), while also improving occupancy IoU (26.0 to 37.2).
The gain is broad across most classes.
Large improvements are observed on both object and structure categories, e.g.
car ($+8.8$), truck ($+8.6$), construction vehicle ($+10.4$), driveable
surface ($+16.7$), sidewalk ($+16.5$), terrain ($+10.8$), and manmade
($+6.8$).
The only class where our score is lower is traffic cone, which remains a
difficult small-object class in coarse voxel grids.
Overall, this indicates that the training-free pipeline provides strong and
well-calibrated semantic occupancy priors even without any downstream training
and no access to target domain images.

\begin{table*}
	\centering
	\caption{Weakly-supervised semantic occupancy prediction performance on 
	Occ3D-nuScenes validation set. IoU per class and mIoU are reported.
	'others' and 'other\_flat' classes are not included in the mIoU calculation.
	* means that the method is from a preprint.}
	\newcommand*\rotext{\multicolumn{1}{R{45}{1em}}}
	\setlength{\tabcolsep}{2.2pt}
	\resizebox{\textwidth}{!}{
		\begin{tabular}{l|cc|cccccc}
			\midrule
			Method                                                                 &  \shortstack{Train\\Free}   & \shortstack{Cam.\\Masks}  & mIoU          & $\text{IoU}_{occ}$ & RayIoU & $1m$ & $2m$ & $4m$ \\
			\midrule

			EasyOcc* \cite{hayes_easyocc_2025}                                     &  $\times$     & $\checkmark$ & 16.0         & 38.9     &      -        &       -              &       -              &      -                \\

			GaussianFlowOcc \cite{boeder_gaussianflowocc_2025}                     &  $\times$     & $\checkmark$ & 17.1         & 13.9              & 16.5         & 11.8                & 16.6                & 21.0                \\
			ShelfOcc* \cite{boeder_shelfocc_2025}                                  &  $\times$     & $\times$     & 9.6          & 26.0               & -             & -                    & -                    & -                    \\

			ShelfOcc* \cite{boeder_shelfocc_2025} (+STCOcc \cite{liao_stcocc_2025})&  $\times$     & $\checkmark$ & \best{22.9}  & \best{56.1}       & \second{20.0} & \best{14.4}         & \second{20.1}       & \second{25.5}                \\

			\midrule
			\method{} (Ours)                                                                   &  $\checkmark$ & $\times$     & 16.9         & 37.2              & 16.5         & 11.7                & 16.7                & 21.6                \\

			\method{} (+STCOcc \cite{liao_stcocc_2025})                                 &  $\times$     & $\times$     & \second{22.8} &  \second{43.8}             & \best{21.1 } &   \second{13.6}     & \best{21.3}         & \best{28.5}                \\

		\end{tabular}
	}
	\label{tab:weak_sup}
\end{table*}

The weakly supervised comparison table~\ref{tab:weak_sup} extends the analysis beyond the strict
train-free setting.
In train-free mode, \method{} reaches 16.9 mIoU and 16.5 RayIoU, which is
very close to GaussianFlowOcc \cite{boeder_gaussianflowocc_2025} in mIoU (17.1), 
on par in overall RayIoU
(16.5 vs.\ 16.5), and slightly better at longer ranges
(RayIoU$_{2m}$: 16.7 vs.\ 16.6, RayIoU$_{4m}$: 21.6 vs.\ 21.0).

When we use our labels for downstream training, \method{}+STCOcc reaches
22.8 mIoU, close to ShelfOcc+STCOcc (22.9), and improves RayIoU from 20.0
to 21.1.
We note that longer range RayIoU metrics are significantly improved
(RayIoU$_{2m}$: 21.3 vs.\ 20.1, RayIoU$_{4m}$: 28.5 vs.\ 25.5),
while the near-range RayIoU$_{1m}$ remains slightly lower
(13.6 vs.\ 14.4).

A key setup difference is that our transfer results do not use camera-visibility
masks during downstream training.
ShelfOcc generates visibility masks for pseudo-labels,
which help to focus the training on visible parts.  
The authors report a large performance drop without these masks 
(13.4 mIoU vs.\ 22.9 mIoU
with masks).
This may explain ShelfOcc's advantage in voxel mIoU, while \method{} dominates
RayIoU, especially for long ranges where there may be more hidden parts.
We claim that training without them may also lead to more reliable results
beyond the benchmark metrics, since masking non-visible parts 
removes all incentives for the model to make correct predictions for those parts
as well.

\begin{table*}
	\centering
	\caption{Panoptic occupancy prediction on Occ3D-nuScenes' validation set. Measured with RayPQ from \cite{liu_fully_2024}.
	}
	\newcommand*\rotext{\multicolumn{1}{R{45}{1em}}}
	\setlength{\tabcolsep}{2.2pt}
	\begin{tabular}{l|cccccc}
		\hline
		Method                                             & Sup.       & RayPQ & $\text{RayPQ}_{1m}$ & $\text{RayPQ}_{2m}$ & $\text{RayPQ}_{4m}$ \\
		\hline
		SparseOcc \cite{liu_fully_2024}                    & Full       & \second{14.1}  & \second{10.2}                & \second{14.5}            & \second{17.6}                \\
		Panoptic-FlashOcc \cite{yu_panoptic-flashocc_2024} & Full       & \best{16.0}  & \best{11.9}                & \best{16.3}            & \best{19.7}                \\
		\hline
		\method{}    (Ours)                                          & Train-free & 3.1   & 0.8                 & 2.6             & 5.8                 \\

		\method{} (+STCOcc \cite{liao_stcocc_2025})             & Weak       & 3.9     & 1.0               & 3.0             & 7.6                   \\

	\end{tabular}
	\label{tab:pano}
\end{table*}

The panoptic results table~\ref{tab:pano} reports panoptic occupancy performance.
In train-free mode, \method{} reaches 3.1~RayPQ; with weak supervision through
STCOcc\_pano, performance increases to 3.9~RayPQ.

Absolute RayPQ remains lower than fully supervised methods, but the range-wise
behavior is informative: RayPQ$_{4m}$ is $103\%$ higher than global RayPQ,
while the same comparison is $23\%$ for Panoptic-FlashOcc,
indicating that geometric alignment quality is a dominant factor in \method{}'s
performance.
This is consistent with the strict RayPQ matching criterion, where correct
panoptic assignment requires both semantic consistency and sufficient occupancy
overlap. True Positives are accounted only if the IoU between the 
predicted segment and the
ground truth segment is above a threshold of 50\%, which is a hard target
for weakly supervised methods even if instances are well detected and 
separated.
We therefore position these numbers as a first training-free/weakly supervised
panoptic baseline and a concrete starting point for future refinements.

\begin{table}
	\small
	\setlength{\tabcolsep}{2pt}
	\centering
	\caption{\textbf{Ablation study.}
		Five stages of ablation are presented where features are added incrementally.
		(I) only includes the Semantic and Geometric branches. (V) is the final
		\method{} method. All values are computed on the Occ3D-nuScenes validation set.}
	\begin{tabular}{c|cccc|ccc}
		\toprule
		Stage & \shortstack{Prompts\\\&Rules} & \shortstack{Conf.\\Filter} & \shortstack{Occ\\Refine} & \shortstack{Inst.\\Ident.} & mIoU & RayIoU & RayPQ \\
		\midrule
		\rowcolor{blue!3}

		(I)                            &        &             &            &                & 10.9 & 11.2  & 1.1  \\
		(II)                           & \cmark &             &            &                & 13.6 & 14.2  & 1.4  \\
		(III)                          & \cmark & \cmark      &            &                & 13.7 & 14.2  & 1.4  \\
		(IV)                           & \cmark & \cmark      & \cmark     &                & 15.8 & 14.6  & 1.5  \\
		\rowcolor{blue!10}
		(V)                            & \cmark & \cmark      & \cmark     & \cmark         & 16.9 & 16.5  & 2.5  \\

		\bottomrule
	\end{tabular}

	\vspace*{2mm}
	\label{tab:ablation}
\end{table}

Table~\ref{tab:ablation} confirms the contribution of each module.
Starting from the semantic+geometric branches (Stage I), prompt/rule semantics
already provide a large gain (Stage II), and occupancy refinement further
improves semantic quality (Stage IV).
Numerically, Stage I $\rightarrow$ II brings +2.7 mIoU, +2.9 RayIoU, and
+0.3 RayPQ, showing that semantic prompt design is a primary driver,
and changes as simple as synonyms may significantly impact performance.
Stage II $\rightarrow$ III is nearly unchanged, indicating that confidence
filtering mostly stabilizes outputs rather than shifting aggregate metrics.
Stage III $\rightarrow$ IV contributes +2.1 mIoU and +0.5 RayIoU through
voxel refinement.
The full model (Stage V) performs best on all reported metrics, and the
instance-identification stage provides the largest panoptic jump
(RayPQ 1.5 $\rightarrow$ 2.5), while also improving RayIoU
(14.6 $\rightarrow$ 16.5).
Overall, Stage I $\rightarrow$ V yields +6.0 mIoU, +5.3 RayIoU, and
+1.4 RayPQ.

\begin{table*}
	\centering
	\caption{Impact of ablating prior extrinsics (no camera poses given to MapAnything as input) 
    and causality (using future frames for current prediction) on panoptic occupancy prediction.
    Values are reported for Occ3D-nuScenes' validation set.}
	\newcommand*\rotext{\multicolumn{1}{R{45}{1em}}}
	\setlength{\tabcolsep}{2.2pt}

		\begin{tabular}{cc|cccccc}
			\hline
			    Extrinsics   &  Causal  & mIoU  & RayIoU & RayPQ \\
			\midrule
                $\checkmark$ & $\checkmark$ & 16.9         & 16.5 & 2.5 \\
                $\times$ & $\checkmark$     & 7.9         & 9.2 & 1.1 \\
                $\checkmark$ & $\times$     & 20.4         & 16.8 & 3.5 \\
            
            \midrule
		\end{tabular}

    \label{tab:ablation_2}
\end{table*}

Table~\ref{tab:ablation_2} provides insights about two other factors:
camera-pose priors and causality constraints.
Removing extrinsics (no camera poses to MapAnything) significantly degrades
performance: mIoU drops by $53\%$, RayIoU by $44\%$, and RayPQ
by $56\%$.
This highlights that while the method is training-free and open-vocabulary,
accurate poses are still important for reliable 3D fusion, and closing this gap
is an interesting direction for future work.
Using future frames (non-causal) improves voxel mIoU by $21\%$ but increases RayIoU
by only $1.8\%$.
This suggests that denser temporal accumulation adds volumetric coverage but 
does not necessarily improve geometric and semantic accuracy beyond
a certain point, given that 
RayIoU is specifically designed to be robust to geometric cheating with
thick surfaces.

\section{Conclusion}
\label{sec:conclusion}

We presented \method{}, a training-free pipeline for camera-only semantic and
panoptic occupancy prediction that combines promptable 2D priors with
foundation-model 3D reconstruction. 
The method can operate without any downstream
training, requires no target data, and preserves 
the prompt-driven flexibility of the open-vocabulary foundation models. 
In this setting, \method{} matches the performance of trained pipelines
on Occ3D-nuScenes, and when used as a pseudo-label generator with minimal 
adaptations, surpasses state-of-the-art methods in RayIoU.
\method{} also sets the
first training-free and weakly supervised panoptic 
occupancy prediction baselines.

Overall, the experiments support three points: 
\begin{enumerate}
    \item Train-free occupancy prediction
is made possible, with \method{}'s performance being competitive with trained
weakly supervised pipelines. 
    \item Panoptic occupancy is feasible without dense 3D supervision.
    \item At the same time, the remaining panoptic gap to fully supervised methods
highlights that geometry quality and precise volumetric alignment remain the
main bottlenecks in this label-free regime.
\end{enumerate}

We hope that future works will explore sensor extrinsics-free performance, which is
currently significantly lower, in order to advance towards even more generalizable
occupancy prediction methods, requiring only camera input.

{
	\small
	\bibliographystyle{splncs04}
	\bibliography{main}

@misc{kirillov_segment_2023,
	title = {Segment {Anything}},
	url = {http://arxiv.org/abs/2304.02643},
	doi = {10.48550/arXiv.2304.02643},
	urldate = {2023-05-22},
	publisher = {arXiv},
	author = {Kirillov, Alexander and Mintun, Eric and Ravi, Nikhila and Mao, Hanzi and Rolland, Chloe and Gustafson, Laura and Xiao, Tete and Whitehead, Spencer and Berg, Alexander C. and Lo, Wan-Yen and Dollár, Piotr and Girshick, Ross},
	month = apr,
	year = {2023},
	note = {arXiv:2304.02643 [cs]},
}

@misc{caesar_nuscenes_2020,
	title = {{nuScenes}: {A} multimodal dataset for autonomous driving},
	shorttitle = {{nuScenes}},
	url = {http://arxiv.org/abs/1903.11027},
	doi = {10.48550/arXiv.1903.11027},
	urldate = {2023-10-06},
	publisher = {arXiv},
	author = {Caesar, Holger and Bankiti, Varun and Lang, Alex H. and Vora, Sourabh and Liong, Venice Erin and Xu, Qiang and Krishnan, Anush and Pan, Yu and Baldan, Giancarlo and Beijbom, Oscar},
	month = may,
	year = {2020},
	note = {arXiv:1903.11027 [cs, stat]},
}

@misc{pan_renderocc_2024,
	title = {{RenderOcc}: {Vision}-{Centric} {3D} {Occupancy} {Prediction} with {2D} {Rendering} {Supervision}},
	shorttitle = {{RenderOcc}},
	url = {http://arxiv.org/abs/2309.09502},
	doi = {10.48550/arXiv.2309.09502},
	urldate = {2025-10-27},
	publisher = {arXiv},
	author = {Pan, Mingjie and Liu, Jiaming and Zhang, Renrui and Huang, Peixiang and Li, Xiaoqi and Wang, Bing and Xie, Hongwei and Liu, Li and Zhang, Shanghang},
	month = mar,
	year = {2024},
	note = {arXiv:2309.09502 [cs]},
}

@misc{tian_occ3d_2023,
	title = {{Occ3D}: {A} {Large}-{Scale} {3D} {Occupancy} {Prediction} {Benchmark} for {Autonomous} {Driving}},
	shorttitle = {{Occ3D}},
	url = {http://arxiv.org/abs/2304.14365},
	doi = {10.48550/arXiv.2304.14365},
	urldate = {2025-10-28},
	publisher = {arXiv},
	author = {Tian, Xiaoyu and Jiang, Tao and Yun, Longfei and Mao, Yucheng and Yang, Huitong and Wang, Yue and Wang, Yilun and Zhao, Hang},
	month = dec,
	year = {2023},
	note = {arXiv:2304.14365 [cs]},
}

@misc{gan_gaussianocc_2025,
	title = {{GaussianOcc}: {Fully} {Self}-supervised and {Efficient} {3D} {Occupancy} {Estimation} with {Gaussian} {Splatting}},
	shorttitle = {{GaussianOcc}},
	url = {http://arxiv.org/abs/2408.11447},
	doi = {10.48550/arXiv.2408.11447},
	urldate = {2025-10-29},
	publisher = {arXiv},
	author = {Gan, Wanshui and Liu, Fang and Xu, Hongbin and Mo, Ningkai and Yokoya, Naoto},
	month = jul,
	year = {2025},
	note = {arXiv:2408.11447 [cs]},
}

@article{huang_tri-perspective_2023,
	title = {Tri-{Perspective} {View} for {Vision}-{Based} {3D} {Semantic} {Occupancy} {Prediction}},
	copyright = {https://doi.org/10.15223/policy-029},
	url = {https://ieeexplore.ieee.org/document/10203437/},
	doi = {10.1109/CVPR52729.2023.00890},
	urldate = {2025-11-07},
	journal = {2023 IEEE/CVF Conference on Computer Vision and Pattern Recognition (CVPR)},
	publisher = {IEEE},
	author = {Huang, Yuanhui and Zheng, Wenzhao and Zhang, Yunpeng and Zhou, Jie and Lu, Jiwen},
	month = jun,
	year = {2023},
	note = {Conference Name: 2023 IEEE/CVF Conference on Computer Vision and Pattern Recognition (CVPR)
ISBN: 9798350301298
Place: Vancouver, BC, Canada},
	pages = {9223--9232},
}

@misc{boeder_gaussianflowocc_2025,
	title = {{GaussianFlowOcc}: {Sparse} and {Weakly} {Supervised} {Occupancy} {Estimation} using {Gaussian} {Splatting} and {Temporal} {Flow}},
	shorttitle = {{GaussianFlowOcc}},
	url = {http://arxiv.org/abs/2502.17288},
	doi = {10.48550/arXiv.2502.17288},
	urldate = {2025-11-07},
	publisher = {arXiv},
	author = {Boeder, Simon and Gigengack, Fabian and Risse, Benjamin},
	month = aug,
	year = {2025},
	note = {arXiv:2502.17288 [cs]},
}

@misc{wei_surroundocc_2023,
	title = {{SurroundOcc}: {Multi}-{Camera} {3D} {Occupancy} {Prediction} for {Autonomous} {Driving}},
	shorttitle = {{SurroundOcc}},
	url = {http://arxiv.org/abs/2303.09551},
	doi = {10.48550/arXiv.2303.09551},
	urldate = {2025-11-07},
	publisher = {arXiv},
	author = {Wei, Yi and Zhao, Linqing and Zheng, Wenzhao and Zhu, Zheng and Zhou, Jie and Lu, Jiwen},
	month = aug,
	year = {2023},
	note = {arXiv:2303.09551 [cs]},
}

@article{keetha_mapanything_nodate,
	title = {{MapAnything}: {Universal} {Feed}-{Forward} {Metric} {3D} {Reconstruction} map-anything.github.io},
	language = {en},
	author = {Keetha, Nikhil and Müller, Norman and Schönberger, Johannes and Porzi, Lorenzo and Zhang, Yuchen and Fischer, Tobias and Knapitsch, Arno and Zauss, Duncan and Weber, Ethan and Antunes, Nelson and Luiten, Jonathon and Lopez-Antequera, Manuel and Bulò, Samuel Rota and Richardt, Christian and Ramanan, Deva and Scherer, Sebastian and Kontschieder, Peter},
}

@misc{liu_fully_2024,
	title = {Fully {Sparse} {3D} {Occupancy} {Prediction}},
	url = {http://arxiv.org/abs/2312.17118},
	doi = {10.48550/arXiv.2312.17118},
	urldate = {2026-02-12},
	publisher = {arXiv},
	author = {Liu, Haisong and Chen, Yang and Wang, Haiguang and Yang, Zetong and Li, Tianyu and Zeng, Jia and Chen, Li and Li, Hongyang and Wang, Limin},
	month = jul,
	year = {2024},
	note = {arXiv:2312.17118 [cs]},
}

@article{wang_opus_nodate,
	title = {{OPUS}: {Occupancy} {Prediction} {Using} a {Sparse} {Set}},
	language = {en},
	author = {Wang, Jiabao and Liu, Zhaojiang and Meng, Qiang and Yan, Liujiang and Wang, Ke and Yang, Jie and Liu, Wei and Hou, Qibin and Cheng, Ming-Ming},
}

@misc{boeder_shelfocc_2025,
	title = {{ShelfOcc}: {Native} {3D} {Supervision} beyond {LiDAR} for {Vision}-{Based} {Occupancy} {Estimation}},
	shorttitle = {{ShelfOcc}},
	url = {http://arxiv.org/abs/2511.15396},
	doi = {10.48550/arXiv.2511.15396},
	urldate = {2026-02-15},
	publisher = {arXiv},
	author = {Boeder, Simon and Gigengack, Fabian and Roesler, Simon and Caesar, Holger and Risse, Benjamin},
	month = nov,
	year = {2025},
	note = {arXiv:2511.15396 [cs]},
}

@misc{yu_panoptic-flashocc_2024,
	title = {Panoptic-{FlashOcc}: {An} {Efficient} {Baseline} to {Marry} {Semantic} {Occupancy} with {Panoptic} via {Instance} {Center}},
	shorttitle = {Panoptic-{FlashOcc}},
	url = {http://arxiv.org/abs/2406.10527},
	doi = {10.48550/arXiv.2406.10527},
	urldate = {2026-02-28},
	publisher = {arXiv},
	author = {Yu, Zichen and Shu, Changyong and Sun, Qianpu and Bian, Yifan and Wei, Xiaobao and Yu, Jiangyong and Liu, Zongdai and Yang, Dawei and Li, Hui and Chen, Yan},
	month = oct,
	year = {2024},
	note = {arXiv:2406.10527 [cs]},
}

@misc{liao_stcocc_2025,
	title = {{STCOcc}: {Sparse} {Spatial}-{Temporal} {Cascade} {Renovation} for {3D} {Occupancy} and {Scene} {Flow} {Prediction}},
	shorttitle = {{STCOcc}},
	url = {http://arxiv.org/abs/2504.19749},
	doi = {10.48550/arXiv.2504.19749},
	urldate = {2026-03-01},
	publisher = {arXiv},
	author = {Liao, Zhimin and Wei, Ping and Chen, Shuaijia and Wang, Haoxuan and Ren, Ziyang},
	month = apr,
	year = {2025},
	note = {arXiv:2504.19749 [cs]},
}

@misc{hayes_easyocc_2025,
	title = {{EasyOcc}: {3D} {Pseudo}-{Label} {Supervision} for {Fully} {Self}-{Supervised} {Semantic} {Occupancy} {Prediction} {Models}},
	shorttitle = {{EasyOcc}},
	url = {http://arxiv.org/abs/2509.26087},
	doi = {10.48550/arXiv.2509.26087},
	urldate = {2026-03-02},
	publisher = {arXiv},
	author = {Hayes, Seamie and Sistu, Ganesh and Eising, Ciarán},
	month = nov,
	year = {2025},
	note = {arXiv:2509.26087 [cs]},
}

@misc{jiang_gausstr_2025,
	title = {{GaussTR}: {Foundation} {Model}-{Aligned} {Gaussian} {Transformer} for {Self}-{Supervised} {3D} {Spatial} {Understanding}},
	shorttitle = {{GaussTR}},
	url = {http://arxiv.org/abs/2412.13193},
	doi = {10.48550/arXiv.2412.13193},
	urldate = {2026-03-02},
	publisher = {arXiv},
	author = {Jiang, Haoyi and Liu, Liu and Cheng, Tianheng and Wang, Xinjie and Lin, Tianwei and Su, Zhizhong and Liu, Wenyu and Wang, Xinggang},
	month = mar,
	year = {2025},
	note = {arXiv:2412.13193 [cs]},
}

@misc{zhang_occnerf_2024,
	title = {{OccNeRF}: {Advancing} {3D} {Occupancy} {Prediction} in {LiDAR}-{Free} {Environments}},
	shorttitle = {{OccNeRF}},
	url = {http://arxiv.org/abs/2312.09243},
	doi = {10.48550/arXiv.2312.09243},
	urldate = {2026-03-02},
	publisher = {arXiv},
	author = {Zhang, Chubin and Yan, Juncheng and Wei, Yi and Li, Jiaxin and Liu, Li and Tang, Yansong and Duan, Yueqi and Lu, Jiwen},
	month = aug,
	year = {2024},
	note = {arXiv:2312.09243 [cs]},
}
}

\end{document}